\pdfoutput=1
\documentclass[sigconf,nonacm]{acmart}

\AtBeginDocument{%
  \providecommand\BibTeX{{%
    \normalfont B\kern-0.5em{\scshape i\kern-0.25em b}\kern-0.8em\TeX}}}

\usepackage[T1]{fontenc}
\usepackage{microtype}
\usepackage{array}
\usepackage{subcaption}
\usepackage{enumitem}

\usepackage{tikz}
\usetikzlibrary{shapes.geometric, positioning, arrows.meta, calc, fit, shapes.misc}

\tikzset{
    nnbase/.style={trapezium, trapezium angle=75, trapezium stretches=true, draw=black!40, fill=blue!5!gray!15, thick, minimum width=4.5cm, minimum height=0.7cm},
    nneventbase/.style={trapezium, trapezium angle=75, trapezium stretches=true, draw=black!40, fill=blue!5!gray!15, thick, minimum width=2.2cm, minimum height=0.4cm},
    nntower/.style={draw=black!50, fill=black!15, thick, minimum width=0.25cm, minimum height=0.4cm},
    nnnewtower/.style={draw=blue!50!gray, fill=blue!30!gray, thick, minimum width=0.25cm, minimum height=0.4cm},
    nnheadxy/.style={draw=blue!50!gray, fill=blue!30!gray, thick, minimum width=0.25cm, minimum height=0.25cm},
    arrow/.style={->, >={stealth[scale=0.8]}, thick, draw=black!40},
    arrownew/.style={->, >={stealth[scale=0.8]}, thick, draw=blue!50!gray},
    lbl/.style={font=\sffamily\scriptsize, text=black!70},
    lblnew/.style={font=\sffamily\scriptsize\bfseries, text=blue!60!gray}
}

\setcopyright{none}
\acmConference[RecSys '26]{20th ACM Conference on Recommender Systems}{September 27--October 02, 2026}{Minneapolis, MN, USA}
\acmBooktitle{20th ACM Conference on Recommender Systems (RecSys '26), September 27--October 02, 2026, Minneapolis, MN, USA}
\acmDOI{10.1145/3773078.3831878}
\acmISBN{979-8-4007-2284-4/2026/09}
\acmYear{2026}
\copyrightyear{2026}

\title[Learned Cross-Task Relationships in Multi-Task Models]{Learned Cross-Task Relationships in Multi-Task Models}
\titlenote{Accepted to the 20th ACM Conference on Recommender Systems (RecSys '26), Minneapolis, MN, USA, September 27--October 02, 2026. This is the author's version of the work. It is posted here for your personal use. Not for redistribution. The definitive Version of Record was published in RecSys '26, \url{https://doi.org/10.1145/3773078.3831878}.}

\author{Victor Zhang}
\email{victorzhang@google.com}
\affiliation{%
  \institution{Google}
  \city{San Bruno}
  \state{CA}
  \country{USA}
}

\author{Yiping Yuan}
\email{yipingyuan@google.com}
\affiliation{%
  \institution{Google}
  \city{San Bruno}
  \state{CA}
  \country{USA}
}

\author{Florian Raudies}
\email{raudies@google.com}
\affiliation{%
  \institution{Google}
  \city{San Bruno}
  \state{CA}
  \country{USA}
}

\author{Bosun Adeoti}
\email{jabos@google.com}
\affiliation{%
  \institution{Google}
  \city{San Bruno}
  \state{CA}
  \country{USA}
}

\author{Brian Y. C. Leung}
\email{mikebrianleung@google.com}
\affiliation{%
  \institution{Google}
  \city{San Bruno}
  \state{CA}
  \country{USA}
}

\author{Sanjay Surendranath Girija}
\email{sanjaysg@google.com}
\affiliation{%
  \institution{Google}
  \city{San Bruno}
  \state{CA}
  \country{USA}
}

\author{Naijing Zhang}
\email{njzhang@google.com}
\affiliation{%
  \institution{Google}
  \city{San Bruno}
  \state{CA}
  \country{USA}
}

\renewcommand{\shortauthors}{Zhang et al.}

\begin{abstract}
We propose a framework that learns cross-task relationships in multi-task models by approximating the joint distribution of task labels through targeted pairwise relationships. This approach improves performance via transfer learning and enhances information extraction without the intractable complexity of modeling the full joint space. Although our framework applies to any multi-task system, we demonstrate its efficacy within YouTube's production recommendation systems. Experiments across the Notifications, Homepage, and Watch Next surfaces show improvements in both accuracy and user satisfaction metrics. Finally, we propose a workflow template to facilitate broader future implementation.
\end{abstract}



\begin{document}

\maketitle

\section{Introduction}
Multi-task learning (MTL) \citep{caruana1993multitask, crawshaw2020multi} posits that simultaneously training on multiple related tasks yields superior performance compared to learning them in isolation. This is typically achieved through a shared neural architecture, as illustrated in Figure~\ref{fig:cross_heads_on_shared_body}, where the shared body is trained using a weighted sum of the individual task losses. By consolidating learning signals into a unified model, the system develops a ``shared representation''---a rich feature space that leverages cross-task relationships to improve both generalization and prediction accuracy, facilitating inductive transfer between related tasks. Furthermore, the shared computational budget allows for efficient resource utilization, enabling the model to scale effectively across tasks.

While multi-task learning has become an industry standard, we identify a fundamental limitation in how its modeling objective is typically framed. Although the shared representation implicitly captures the relationships between tasks (by training on multiple labels simultaneously), this rich relational information is rarely explicitly quantified as a model output. To illustrate this, consider an object detection model trained to identify various entities, such as a cat and a pet food bowl. While the model's internal representation may learn that these objects are highly correlated, the final output is typically restricted to the independent (marginal) probabilities of each item, $P(\text{cat})$ and $P(\text{pet bowl})$, discarding the joint structural information. In the language of probability, the training data labels are sampled from the \textit{joint distribution}, $(y_1, \dots, y_N) \sim \mathcal{D}(Y_1, \dots, Y_N)$, of the $N$ tasks, whereas the model may output only a vector of mean predictions, $(P_1, \dots, P_N)$, estimating the $N$ \textit{marginal distributions}. Consequently, while the label space possesses exponential underlying informational complexity in $N$, the standard model output is significantly reduced to linear complexity in $N$. We elaborate on this formulation in Section~\ref{Mathematical Formulation}.

This limitation suggests that multi-task models are constrained in their expressiveness, failing to fully describe the cross-task relationships regardless of the computational and engineering efforts to improve the accuracy of individual task predictions.

In this paper, we make the following contributions:
\begin{enumerate}
    \item We formalize the concept of learning cross-task relationships as an approximation of the joint distribution via pairwise task interactions (e.g., covariance), and demonstrate how this can be efficiently implemented by strategically adding auxiliary heads to an existing multi-task model.
    \item We propose two primary hypotheses explaining how learned task relationships enhance the performance of downstream reinforcement learning (RL) systems that utilize multi-task predictions, and we detail the implementation and testing workflow for each.
    \item We describe the implementation and results from full deployments and experiments across multiple YouTube recommender systems---specifically Notifications, Homepage, and the Watch Next sidebar---where multi-task learning is utilized to characterize user behavior through diverse content interaction tasks.
\end{enumerate}

\subsection{Related Work}

The use of auxiliary heads to enhance transfer learning in MTL is an established technique in large-scale recommender models \citep{jaderberg2016reinforcement, ruder2017overview}. However, standard auxiliary-head frameworks typically focus on predicting independent, marginal probabilities of supplementary tasks to implicitly enrich the shared representation instead of extracting information on cross-task relations.

Our novelty lies in the strategic selection of pairwise targets. Rather than adding arbitrary related tasks, our approach---detailed in Section~\ref{Deriving Covariance}---explicitly constructs auxiliary heads to capture structural cross-task relationships (e.g., cross-labels). By framing these specific auxiliary heads as estimators of covariance, our method provides a targeted mathematical approximation of the joint distribution. This aligns with existing research on characterizing uncertainty in deep learning, such as estimating predictive variance \citep{kendall2017uncertainties}, but applies the concept to inter-task dependencies.

Furthermore, prior work \citep{dabney2018distributional} has shown that for regression tasks, learning a distribution over quantiles via multiple heads and subsequently deriving the mean yields superior accuracy compared to direct mean estimation---a methodology that has also been deployed in production at YouTube. This parallels our objective of learning a more complete probability model; however, rather than estimating the full distribution of a single task, we focus on estimating the joint distribution across multiple tasks.

Well-studied architectures such as MMoE \citep{ma2018modeling} and PLE \citep{tang2020progressive} improve the shared representation via gating mechanisms, but unlike our approach, they still rely on independent marginal losses. To address the inherent limitations of assuming task independence, a parallel line of research explicitly models structural relationships between tasks and labels. For instance, \citet{cao2022relational} propose a relational multi-task learning framework that explicitly quantifies dependencies between data points and tasks. Similarly, in the multi-label classification domain, \citet{chen2019multi} utilize Graph Convolutional Networks to map out label correlations, while \citet{yeh2017learning} derive a deep latent space specifically optimized for label co-occurrence. These works share our motivation that discarding joint structural information inherently limits model expressiveness; however, they primarily focus on architectural routing or graph-based mapping rather than targeted mathematical approximations of covariance via auxiliary heads.

Within the specific context of recommender systems, handling nested or sequentially dependent tasks (e.g., an impression leading to a click, which in turn leads to a downstream action like a save or inline playback) is a well-recognized challenge. The Entire Space Multi-Task Model (ESMM) \citep{ma2018entire} addresses sample selection bias by estimating the joint probability over the entire space utilizing the exact probability chain rule---$P(\text{click} \cap \text{conversion}) = P(\text{conversion} \mid \text{click})P(\text{click})$---that underpins our derivation for conditional tasks in Section~\ref{Deriving Covariance}. Building upon this, \citet{xi2021modeling} propose the Adaptive Information Transfer Multi-task (AITM) model to explicitly capture sequential dependence among multi-step user conversions. While these foundational works utilize conditional probability structures to model deep user journeys, our novelty lies in introducing targeted auxiliary unconditional and cross-label heads. This uniquely forces the shared representation to encode discriminative features specific to the interaction of tasks, deliberately optimizing the environmental state space for downstream reinforcement learning policies.

A comprehensive survey by \citet{crawshaw2020multi} reviews approaches to task relationship learning. These methods include task grouping strategies, identifying which related tasks maximize transfer for a target task, and computing quantitative measures of inter-task similarity.

Our methodology also offers a novel structural perspective on the well-documented challenge of task gradient conflict in multi-task learning \citep{yu2020gradient}. When optimizing a shared representation across independent marginal tasks, gradients often point in conflicting directions or vary wildly in magnitude, leading to destructive interference and suboptimal convergence. A rich body of optimization literature has addressed this by intervening during backpropagation: altering gradient magnitudes \citep{chen2018gradnorm}, projecting conflicting gradients via gradient surgeries \citep{yu2020gradient}, or enforcing conflict-averse descent trajectories \citep{liu2021conflict}. Instead of manipulating the gradients algorithmically after they are computed, our approach structurally prevents gradient dilution. By introducing a cross-task head, the model explicitly generates a unified gradient signal perfectly aligned with the intersection of both tasks. We hypothesize that this auxiliary loss structurally strengthens the gradient magnitudes for overlapping features, guiding the shared representation toward a cooperative minima without the need for complex, dynamic gradient routing or weighting schemes.

\subsection{Multi-Task Learning in YouTube Recommender Systems}

\begin{figure}[htbp]
    \centering
    \resizebox{0.8\columnwidth}{!}{%
    \begin{tikzpicture}[
        node distance=0pt,
        box/.style={draw=black!70, thick, align=center, inner sep=2.5pt},
        intro/.style={box, fill=red!15},
        backprop/.style={box, fill=orange!20},
        transfer/.style={box, fill=cyan!15},
        unchanged/.style={box, fill=gray!20},
        input/.style={box, minimum height=0.55cm, font=\sffamily\footnotesize},
        tower/.style={box, minimum height=0.8cm, minimum width=1.1cm, font=\sffamily\scriptsize},
        mainarrow/.style={->, >=stealth, very thick, draw=black!80},
        phase2arrow/.style={->, >=stealth, line width=1.8pt, draw=black!80}
    ]

    \def\fullwidth{7.8cm}
    \def\halfwidth{3.9cm}
    \def\quarterwidth{1.95cm}

    \node[unchanged, minimum width=\fullwidth, minimum height=0.7cm, font=\sffamily\normalsize\bfseries] (lrf body) at (0,0) {Learned Policy Function};
    
    \node[transfer, input, minimum width=\halfwidth, below=0pt of lrf body.south west, anchor=north west] (lrf in1) {Engagement Scores};
    \node[unchanged, input, minimum width=\halfwidth, right=0pt of lrf in1] (lrf in2) {Other Input Features};

    \node[font=\sffamily\small, align=center] (exist text) at (-1.9, -2.4) {Existing Heads};
    \node[font=\sffamily\small\bfseries, align=center] (new text) at (1.9, -2.3) {Task Relation Heads\\[-0.2ex]\normalfont (Phase 1, Sec.~\ref{Transfer Learning Hypothesis})};

    \node[transfer, tower, below=0.15cm of exist text] (t2) {Visit};
    \node[transfer, tower, left=0.1cm of t2] (t1) {Click};
    \node[transfer, tower, right=0.1cm of t2] (t3) {Save};

    \node[intro, tower, below=0.15cm of new text, xshift=-0.6cm] (t4) {Save\\Uncond.};
    \node[intro, tower, below=0.15cm of new text, xshift=0.6cm] (t5) {Save $\times$\\Share};

    \node[draw=black!70, dashed, thick, inner sep=4pt, fit=(t1) (t3) (exist text)] (exist group) {};
    \node[draw=black!70, dashed, thick, inner sep=4pt, fit=(t4) (t5) (new text)] (new group) {};

    \draw[mainarrow] (t1.north |- exist group.north) -- (t1.north |- lrf in1.south);
    
    \draw[phase2arrow, dotted, rounded corners=2pt, >={stealth[scale=1.2]}] 
        (new group.north) 
        -- ++(0, 0.45) coordinate (turn) 
        -- (turn -| -1.85,0) 
        node[anchor=west, fill=white, inner sep=2pt, font=\sffamily\small\bfseries] (p2label) {New Input \normalfont(Phase 2, Sec.~\ref{Environment Modeling Hypothesis})}
        (p2label.west) 
        -| (t2.north |- lrf in1.south);

    \path (exist group.south) -- (new group.south) coordinate[midway] (mid_groups);
    \node[backprop, minimum width=\fullwidth, minimum height=0.6cm, below=0.25cm of mid_groups, font=\sffamily\small] (shared) {Shared Body};

    \draw[mainarrow, >={stealth[scale=1.2]}] (shared.north -| exist group.south) -- (exist group.south);
    \draw[mainarrow, >={stealth[scale=1.2]}] (shared.north -| new group.south) -- (new group.south);

    \node[backprop, minimum width=\fullwidth, minimum height=0.7cm, below=0pt of shared, font=\sffamily\normalsize\bfseries] (esm body) {Engagement Scoring Model (MTL)};

    \node[unchanged, input, minimum width=\fullwidth, below=0pt of esm body] (esm in) {Input Features};

    \node[unchanged, minimum width=\fullwidth, minimum height=0.7cm, below=0.3cm of esm in.south, font=\sffamily\normalsize] (retrieval) {Content Retrieval};
    \draw[mainarrow] (retrieval.north) -- (esm in.south);

    \end{tikzpicture}
    }
    \caption{Recommender System Architecture.}
    \Description{General Architecture of Adding Heads}
    \label{fig:recsys_architecture}
\end{figure}
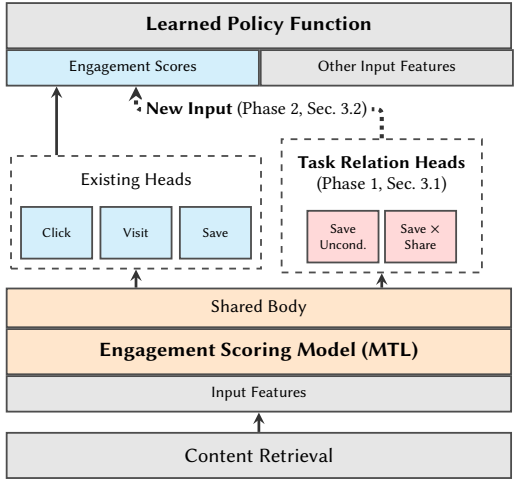

The following two-step process, illustrated in Figure~\ref{fig:recsys_architecture}, demonstrates how YouTube recommender systems integrate multi-task learning within an RL framework and establishes the terminology used throughout this paper.

\begin{enumerate}[leftmargin=*]
    \item \textbf{State Representation via MTL:} A learned representation of user behavior is generated by a large, complex multi-task model, which we call the \textit{User Engagement Model}. The output of this model forms a critical part of the state space $\mathcal{S}$ in the standard RL formulation \citep{sutton1998reinforcement}. The User Engagement Model predicts distinct responses to content---such as the probability of a click, save, or dismissal---as separate tasks. Key modeling considerations for this architecture are described in \citep{covington2016deep} and \citep{zhao2019recommending}.

    \item \textbf{Policy Execution:} These predictions serve as input features for a \textit{decision policy}, derived from a reinforcement learning algorithm \citep{chen2019topk, wu2024learned}, which selects an action from the action space $\mathcal{A}$ (e.g., whether or not to send a notification) that maximizes a reward objective. The policy may incorporate a learned neural network component---as described in \citep{wu2024learned}---which is typically smaller than the User Engagement Model. We refer to such components as the \textit{learned policy functions}.
\end{enumerate}

\section{Mathematical Formulation}\label{Mathematical Formulation}

\subsection{Marginal and Joint Distributions}\label{Marginal and Joint Distributions}

Consider a multi-task learning scenario where a user's actions in response to a piece of content are modeled in a multi-task model, e.g., through predicting the probability of a click, save, or like, and the expected watch time. These separate events can be considered as random variables whose probability (or expected value) the multi-task model predicts. The output of the multi-task model is classically a collection of separate predictions, one for each task, e.g., $P(\text{click}), P(\text{save}), \mathbb{E}[\text{watch\_time}]$.

However, the collection of separate per-task predictions does not constitute a complete model of the user. The full complexity---assuming we take these actions (click, save, etc.) as the fundamental ground truth---is contained in the joint distribution between the tasks. In the language of probability, the separate prediction tasks output the marginal distribution for each task, which is obtained from the joint distribution by integrating out the random variables of the other tasks, a process that effectively destroys information.

Consider the following illustrative example of the loss of information using two actions: click and save. The marginal distributions may be defined by just \textit{two} values: $P(\text{click})=0.2$, $P(\text{save})=0.1$. 
In contrast, the joint distribution is defined by Table~\ref{tab:click_save}, which has \textit{three} degrees of freedom. The marginal probabilities are obtained by summing the joint probabilities along the Click and Save axes. \[ P(\text{click})= P(\text{click} \cap \text{save}) + P(\text{click} \cap \text{no\_save}) = 0.05 + 0.15 \]
\[ P(\text{save})=P(\text{click} \cap \text{save})+P(\text{no\_click} \cap \text{save})=0.05+0.05 \]

\begin{table}[htpb]
    \caption{Joint Probability Distribution of Click vs. Save.}
    \label{tab:click_save}
    \centering
    \begin{tabular}{@{} l c c @{}}
        \toprule
         & \textbf{Save} & \textbf{No Save} \\
        \midrule
        \textbf{Click} & 0.05 & 0.15 \\
        \textbf{No Click} & 0.05 & 0.75 \\
        \bottomrule
    \end{tabular}
\end{table}

To rigorously distinguish our approach from standard multi-task learning, consider two tasks $X$ and $Y$ with binary ground truth labels $\mathsf{y}_x, \mathsf{y}_y \in \{0,1\}$ and predicted probabilities $\hat{p}_x, \hat{p}_y$. In a standard multi-task setup, the model minimizes a weighted sum of the marginal losses \citep{ma2018modeling}. Assuming binary cross-entropy (BCE) for classification tasks \citep{tang2020progressive}:

\begin{equation}
\mathcal{L}_{\text{standard}} = \lambda_x \mathcal{L}_{\text{BCE}}(\mathsf{y}_x, \hat{p}_x) + \lambda_y \mathcal{L}_{\text{BCE}}(\mathsf{y}_y, \hat{p}_y)
\end{equation}

where $\lambda$ represents the scalar loss weight for each task. Under this formulation, the shared representation $\theta$ is optimized to discriminate the marginal distributions $P(X)$ and $P(Y)$ independently. The gradient signal $\nabla_\theta \mathcal{L}_{\text{standard}}$ does not explicitly penalize the model for failing to capture the interaction between $X$ and $Y$.

Our proposed framework has an option (explained in more detail in Section~\ref{Deriving Covariance}) that introduces an auxiliary head that predicts the joint event $XY$, with ground truth $\mathsf{y}_{xy} = \mathsf{y}_x \cdot \mathsf{y}_y$ (logical AND) and predicted probability $\hat{p}_{xy}$. The total objective becomes:

\begin{equation}
\mathcal{L}_{\text{total}} = \mathcal{L}_{\text{standard}} + \lambda_{xy} \mathcal{L}_{\text{BCE}}(\mathsf{y}_{xy}, \hat{p}_{xy})
\end{equation}

One may argue that if the shared embedding layer is sufficiently robust, the joint information should already be encoded in the latent state, even if the heads are marginal. While shared embeddings implicitly allow for transfer, the standard loss $\mathcal{L}_X + \mathcal{L}_Y$ does not strictly penalize the model for failing to capture the covariance structure, whereas $\mathcal{L}_{xy}$ does.

The theoretical contribution of this additional term is twofold:
\begin{itemize}
    \item \textbf{Covariance Disentanglement:} The cross-task head effectively estimates $\mathbb{E}[XY]$. Since the covariance is defined as $\text{Cov}(X,Y) = \mathbb{E}[XY] - \mathbb{E}[X]\mathbb{E}[Y]$, and the marginal heads already estimate $\mathbb{E}[X]$ and $\mathbb{E}[Y]$, the system collectively learns the necessary components to resolve the correlation structure between tasks; this is explained in Section~\ref{Deriving Covariance}.

    \item \textbf{Gradient Targeting:} Crucially, the gradient $\nabla_{\theta} \mathcal{L}_{xy}$ provides a learning signal specifically for the subset of the data where tasks co-occur. Unlike simple loss weighting, which scales gradients globally, this term forces the shared representation $\theta$ to encode discriminative features specific to the interaction of tasks (e.g., distinguishing a ``long watch'' from a ``long watch + like''), thereby enriching the shared latent space for all downstream tasks.
\end{itemize}
\subsection{Modeling: Approximating Pairwise Relationships}\label{Modeling: Estimates of the Joint Distribution}

Because modeling the full joint distribution of $N$ binary classification tasks requires $O(2^N)$ additional prediction heads, doing so is computationally intractable. Instead, we propose a finely controllable approximation. We strategically add auxiliary heads based on specific pairs of existing heads, so that the labels are crosses of existing labels or unconditioned versions of an existing task. This approach ensures the model captures pairwise cross-task relationships requiring at most $O(N^2)$ heads, without altering the existing label definitions. For example:
\begin{itemize}
    \item If the model's existing save and click heads predict the direct probabilities $P(\text{click})$ and $P(\text{save})$ among many other heads, one of the additional heads is the head predicting the cross-label $P(\text{save} \cap \text{click})$.
    \item In some cases, the existing heads are slightly different: a base prediction of central importance is made, such as $P(\text{click})$, and many other prediction tasks are conditioned on it, such as $P(\text{save} \mid \text{click})$. This is achieved by masking out training examples where the primary label (e.g., click) is negative. In this scenario, one of the added auxiliary heads is the unconditional prediction, $P(\text{save})$, trained on all examples.
\end{itemize}
We provide two explanations for this proposed selection below.

\subsubsection{Deriving Covariance}\label{Deriving Covariance}

Rather than modeling the full joint space, we can capture fundamental cross-task relationships by estimating their covariance. For two random variables $X$ and $Y$, the covariance is:

\[ \mathrm{Cov}(X,Y) = \mathbb{E}[XY] - \mathbb{E}[X]\mathbb{E}[Y] \] The idea is to collect predictions for each pair of tasks $X, Y$ covering all three terms $\mathbb{E}[X], \mathbb{E}[Y], \mathbb{E}[XY]$ that are required to compute the covariance. Typically two are present and one is missing.

\begin{itemize}

\item \textbf{Direct Tasks:} If the model predicts unconditional tasks $X$ and $Y$ (e.g., $X=\text{click}$, $Y=\text{save}$), it already estimates $\mathbb{E}[X]$ and $\mathbb{E}[Y]$. We add an auxiliary head for the cross-label, estimating $\mathbb{E}[XY]$, which in this example is also $P(X \cap Y)$.

\item \textbf{Conditional Tasks:} If the model predicts a binary task $X$ and a conditional task $Y \mid X$, we can derive the joint expectation via $\mathbb{E}[XY] = \mathbb{E}[Y \mid X] \mathbb{E}[X]$. The missing term is $\mathbb{E}[Y]$, and we add an auxiliary \textit{unconditioned} head for $Y$.

\end{itemize}

\subsubsection{Deriving the Pairwise Joint Distribution}

For any binary task pair $X, Y \in \{0, 1\}$, capturing the complete pairwise joint distribution requires modeling all four realization states: $\mathbb{E}[XY]$, $\mathbb{E}[X(\neg Y)]$, $\mathbb{E}[(\neg X)Y]$, and $\mathbb{E}[(\neg X)(\neg Y)]$. 

Because $\neg X = 1-X$ and $\neg Y = 1-Y$, the linearity of expectation guarantees that all four states can be analytically derived using strictly three terms: $\mathbb{E}[X]$, $\mathbb{E}[Y]$, and $\mathbb{E}[XY]$. For example:
\begin{align*}
    \mathbb{E}[X(\neg Y)] &= \mathbb{E}[X(1-Y)] = \mathbb{E}[X] - \mathbb{E}[XY] \\
    \mathbb{E}[(\neg X)(\neg Y)] &= \mathbb{E}[(1-X)(1-Y)] = 1 - \mathbb{E}[X] - \mathbb{E}[Y] + \mathbb{E}[XY]
\end{align*}

Therefore, introducing auxiliary heads to ensure the model outputs these three base components for a given pair is mathematically equivalent to estimating their complete pairwise joint distribution. This is a first-order approximation of the full joint space, which has intractable $O(2^N)$ complexity.

\section{Two Primary Hypotheses}

We propose two fundamental hypotheses for improving multi-task learning by modeling cross-task relationships. While we theoretically ground these hypotheses in joint distribution estimation and targeted gradient signals (Section~\ref{Mathematical Formulation}), our primary validation is empirical through offline model evaluation and live user experiments.

\subsection{Transfer Learning Hypothesis} \label{Transfer Learning Hypothesis}

We hypothesize that adding auxiliary heads to learn cross-task relations improves the performance of existing heads within the multi-task model. We posit that this is because the shared body of the multi-task model will improve when tasked with approximating the joint distribution, and that these benefits will transfer to the existing heads via the shared representation.

This can be verified in both offline and online experiments:
\begin{enumerate}
    \item Offline, by examining the AUC scores of the existing heads before and after the auxiliary cross-task relationship heads are added.
    \item In live user experiments, by serving the multi-task model with the added auxiliary heads but ignoring their outputs, purely to test the empirical impact of the improved performance of the existing heads.
\end{enumerate}

\subsection{Environment Modeling Hypothesis} \label{Environment Modeling Hypothesis}

We hypothesize that there will be improved performance in the RL agent that consumes the multi-task model signals when we incorporate the additional outputs from the heads approximating cross-task relationships.

We attribute this to the fact that the model of the environment---represented by the multi-task model outputs---becomes richer and more informative. In the context of YouTube recommendations, the main way we assess this improvement in the RL agent is by measuring the impact on metrics in A/B tests on live users. For example, it may be useful to know that for a certain class of videos, a click is anti-correlated with a save because the user wants to watch it only once, while for other videos they are correlated because the user finds value in repeating the experience.

\section{Model Architecture}

For the RL agent model, we incorporated the outputs of the additional cross-task relationship heads as standard input features. For the multi-task model, we evaluated two architectural approaches, both of which proved effective. We initially explored the more complex second approach but ultimately adopted the first, which offered greater ease of implementation.

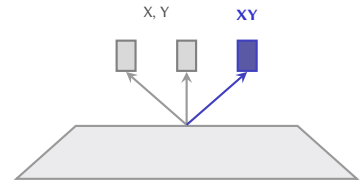
\begin{figure}[htpb]
    \centering
    \begin{tikzpicture}
        \node[nnbase] (base) {};
        \node[nntower, above=0.7cm of base, xshift=-0.8cm] (t1) {};
        \node[nntower, above=0.7cm of base, xshift=0cm] (t2) {};
        \node[nnnewtower, above=0.7cm of base, xshift=0.8cm] (txy) {};
        
        \draw[arrow] (base.north) -- (t1.south);
        \draw[arrow] (base.north) -- (t2.south);
        \draw[arrownew] (base.north) -- (txy.south);
        
        \path (t1.north) -- (t2.north) node[midway, lbl, above=0.15cm] {X, Y};
        \node[lblnew, above=0.15cm of txy] {XY};
    \end{tikzpicture}
    \caption{Connect Directly to Shared Body.}
    \label{fig:cross_heads_on_shared_body}
    \Description{Neural network diagram of added heads connecting directly to shared body.}
\end{figure}

\begin{figure}[htpb]
    \centering
    \begin{tikzpicture}
        \node[nnbase] (base) {};
        \node[nntower, above=0.8cm of base, xshift=-0.5cm] (t1) {};
        \node[nntower, above=0.8cm of base, xshift=0.5cm] (t2) {};
        \node[nnheadxy, above=0.8cm of base, yshift=1.6cm] (hxy) {};
        
        \draw[arrow] (base.north) -- (t1.south);
        \draw[arrow] (base.north) -- (t2.south);
        \draw[arrownew] (t1.north) -- (hxy.south west);
        \draw[arrownew] (t2.north) -- (hxy.south east);
        
        \node[lbl, left=0.15cm of t1] {X};
        \node[lbl, right=0.15cm of t2] {Y};
        \node[lblnew, right=0.15cm of hxy] {XY};
    \end{tikzpicture}
    \caption{Connect to Existing Heads' Top Hidden Layer.}
    \label{fig:cross_heads_through_existing}
    \Description{Diagram showing an auxiliary head connecting directly to the top hidden layers.}
\end{figure}
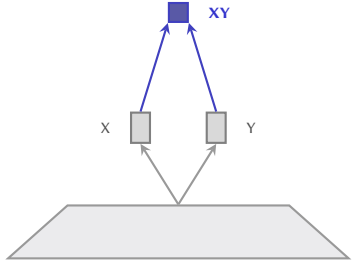

\begin{enumerate}
    \item Directly connect the added head to the shared body without stopping gradients as in Figure~\ref{fig:cross_heads_on_shared_body}. This approach is the simplest, most scalable, and generally preferred based on our experience implementing across multiple YouTube models.

    \item For a cross-label head $XY$ associated with existing heads $X$ and $Y$, we implemented a smaller head (with fewer parameters) that connects directly to the top hidden layers of the specific heads for $X$ and $Y$ without stopping gradients as in Figure~\ref{fig:cross_heads_through_existing}. The rationale is that the representations most useful for learning $XY$ are likely already contained within the specialized representations of $X$ and $Y$. While this approach adds fewer total parameters than connecting to the shared body, the savings are likely negligible relative to the total model size.

\end{enumerate}


\section{Implementation and Testing Workflow}

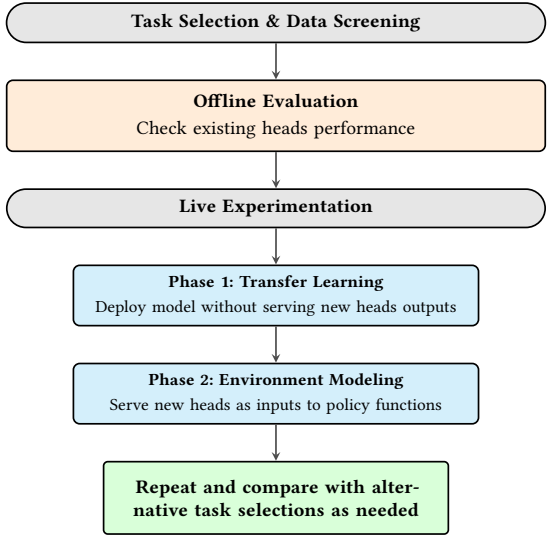
\begin{figure}[htbp]
    \centering
    \resizebox{0.85\columnwidth}{!}{%
    \begin{tikzpicture}[
        node distance=0.6cm and 0.8cm, 
        font=\small,                    
        decision_rect/.style={draw, rectangle, rounded corners=3pt, thick, fill=orange!15, text width=6.0cm, align=center, inner sep=6pt, font=\normalsize\bfseries},
        process/.style={draw, rectangle, rounded corners=3pt, thick, fill=cyan!15, text width=4.5cm, align=center, minimum height=1.0cm},
        action_2line/.style={draw, rectangle, thick, fill=green!15, text width=3.8cm, align=center, minimum height=1.2cm, rounded corners=2pt, font=\small\bfseries},
        terminal/.style={draw, rounded rectangle, thick, fill=gray!20, text width=4.0cm, align=center, inner sep=5pt, font=\normalsize\bfseries},
        arrow/.style={->, >=stealth, thick, draw=black!70},
        label_style/.style={font=\scriptsize\bfseries, fill=white, inner sep=1pt}
    ]

    \node[terminal, text width=8.5cm] (start) {Task Selection \& Data Screening};
    
    \node[decision_rect, below=0.6cm of start, text width=8.5cm] (offline) {Offline Evaluation\\[0.5ex]
        \normalfont Check existing heads performance};
    
    \node[terminal, below=0.6cm of offline, text width=8.5cm] (live) {Live Experimentation};
    
    \node[process, below=0.6cm of live, text width=6.5cm] (phase1) {\textbf{Phase 1: Transfer Learning}\\[0.5ex] \normalfont Deploy model without serving new heads outputs};
    
    \node[process, below=0.6cm of phase1, text width=6.5cm] (phase2) {\textbf{Phase 2: Environment Modeling}\\[0.5ex] \normalfont Serve new heads as inputs to policy functions};

    \node[action_2line, below=0.6cm of phase2.south, text width=5.5cm, font=\bfseries] (end_repeat) {Repeat and compare with alternative task selections as needed};

    \draw[arrow] (start) -- (offline);
    \draw[arrow] (offline) -- (live);
    \draw[arrow] (live) -- (phase1);
    \draw[arrow] (phase1) -- (phase2);
    
    \draw[arrow] (phase2) -- (end_repeat);

    \end{tikzpicture}
    }
    \caption{Experimentation Workflow.}
    \Description{Description of the general workflow of iterating on added cross task correlation heads.}
    \label{fig:experimentation_workflow}
\end{figure}

The following workflow (also illustrated in Figure~\ref{fig:experimentation_workflow}) suggests a way to adopt our contribution in any stack involving a multi-task model and has been used widely across several YouTube surfaces.

\begin{figure}[htbp]
    \centering
    \resizebox{0.85\columnwidth}{!}{%
    \begin{tikzpicture}[
        node distance=0.6cm and 0.4cm, 
        decision_rect/.style={draw, rectangle, rounded corners=3pt, thick, fill=orange!15, text width=6.0cm, align=center, inner sep=4pt, font=\small\bfseries},
        process/.style={draw, rectangle, rounded corners=3pt, thick, fill=cyan!15, text width=3.8cm, align=center, minimum height=0.8cm, inner sep=4pt, font=\small},
        action_2line/.style={draw, rectangle, thick, fill=green!15, text width=3.8cm, align=center, minimum height=0.9cm, rounded corners=2pt, inner sep=4pt, font=\small\bfseries},
        action_side/.style={draw, rectangle, thick, fill=green!15, text width=1.4cm, align=center, minimum height=0.8cm, rounded corners=2pt, inner sep=4pt, font=\small\bfseries},
        terminal/.style={draw, rounded rectangle, thick, fill=gray!20, text width=4.0cm, align=center, inner sep=4pt, font=\small\bfseries},
        arrow/.style={->, >=stealth, thick, draw=black!70},
        label_style/.style={font=\scriptsize\bfseries, fill=white, inner sep=1pt}
    ]

    \node[terminal] (start) {Select Task Pair $(X, Y)$};
    
    \node[decision_rect, below=0.6cm of start] (cond1) {Are $X$ and $Y$ mutually exclusive?};
    \node[decision_rect, below=0.6cm of cond1] (cond2) {Is $Y$ nested within $X$?};
    \node[decision_rect, below=0.6cm of cond2] (cond3) {What are the existing task representations?};
    
    \node[action_side, right=0.6cm of start] (discard) {Discard};
    
    \node[process, below=0.7cm of cond3, xshift=-2.1cm] (direct_tasks) {Both are Unconditioned};
    \node[action_2line, below=0.3cm of direct_tasks] (add_cross) {Add Cross-Label Head:\\[0.5ex]$\mathbb{E}[XY]$};
    
    \node[process, below=0.7cm of cond3, xshift=2.1cm] (cond_tasks) {$Y$ is conditioned on $X$};
    \node[action_2line, below=0.3cm of cond_tasks] (add_uncond) {Add Unconditioned Head:\\[0.5ex]$\mathbb{E}[Y]$};

    \draw[arrow] (start) -- (cond1);
    \draw[arrow] (cond1.south) -- node[left, label_style] {No} (cond2.north);
    \draw[arrow] (cond2.south) -- node[left, label_style] {No} (cond3.north);
    
    \draw[thick, draw=black!70] (cond2.east) -- node[above, label_style, pos=0.6] {Yes} (cond2.east -| discard.south);
    \draw[thick, draw=black!70] (cond1.east) -- node[above, label_style, pos=0.6] {Yes} (cond1.east -| discard.south);
    
    \draw[arrow] (cond2.east -| discard.south) -- (discard.south);
    
    \draw[arrow] (cond3.south -| direct_tasks.north) -- (direct_tasks.north);
    \draw[arrow] (direct_tasks) -- (add_cross);
    
    \draw[arrow] (cond3.south -| cond_tasks.north) -- (cond_tasks.north);
    \draw[arrow] (cond_tasks) -- (add_uncond);

    \end{tikzpicture}
    }
    \caption{Decision Logic for Task Selection.}
    \Description{Diagram to support logic for choosing heads to add.}
    \label{fig:task_selection_logic}
\end{figure}
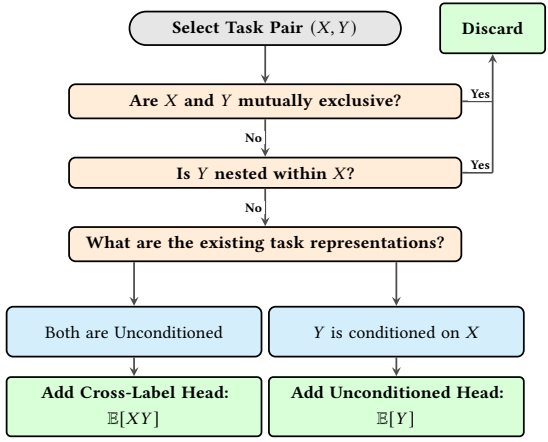

\begin{enumerate}
    \item \textbf{Task Selection and Data Screening:} Select the task pairs for which to estimate cross-task relationships, explicitly excluding pairs with trivial joint distributions. Examples below are also illustrated in Figure~\ref{fig:task_selection_logic}. Additionally, check for data integrity, e.g., $XY$ may be sparse compared to $X$ or $Y$.
    \begin{itemize}
        \item \textbf{Nested Tasks:} For pairs such as $X=\text{open}$ and $Y=\text{open (variant)}$, where $Y$ implies $X$ ($Y \subseteq X$), the cross-task label simplifies to the narrower task: $XY=Y$.
        \item \textbf{Mutually Exclusive Tasks:} For pairs such as $X=\text{open}$ and $Y=\text{dismiss}$, the two actions cannot occur simultaneously, so the joint label is always zero ($XY=0$).
        \item \textbf{Redundant Unconditionals:} For a primary task $X$ (e.g., \textit{click}) and a conditional task $Y \mid X$ where $Y \subseteq X$ (e.g., \textit{watch from click}), the unconditional probability is fully defined by the chain rule: $P(Y) = P(Y \cap X) = P(Y \mid X)P(X)$. Since $P(Y)$ is analytically derivable from the existing predictions, adding it as an explicit head yields no new information. We checked the backend label implementation to make sure the added unconditionals are not redundant.
    \end{itemize}
    \item \textbf{Offline Evaluation:} Add the auxiliary cross-task relationship heads to the multi-task model. Check for AUC gains in the existing tasks, specifically focusing on the original two tasks involved in each cross-task relation.
    \item \textbf{Live Experimentation:} Execute the following two-phase live experiment strategy (illustrated in Figure~\ref{fig:recsys_architecture}):
    \begin{itemize}
        \item \textbf{Phase 1 (Transfer Learning Evaluation):} Deploy the updated multi-task model with the newly added cross-task heads, but \textit{suppress} the downstream consumption of their outputs. This measures the impact of transfer learning to the existing heads in isolation, directly testing the Transfer Learning Hypothesis (Section~\ref{Transfer Learning Hypothesis}).
        \item \textbf{Phase 2 (Environment Modeling Evaluation):} Update the deployment to serve the new cross-task predictions as input features to the Learned Policy Function (LPF). By using the Phase 1 arm as a baseline, this isolates the specific value of the increased information density---validating the Environment Modeling Hypothesis (Section~\ref{Environment Modeling Hypothesis})---distinct from the accuracy gains achieved via transfer learning.
    \end{itemize}
\item \textbf{Iterative Refinement:} Repeat this process with alternative head selections as necessary. Empirical results in a live environment may be difficult to predict, requiring iteration to identify the set of heads that yields a substantial impact on user experience.
\end{enumerate}
\section{Description and Results on YouTube Surfaces}

This section details a cumulative series of launches and experiments conducted across YouTube surfaces, noting unique challenges in contexts where they proved significant. Additional evidence is pending from experiments and deployments that are currently in progress.

\textbf{Validation Procedure and Terminology:} A \textit{live experiment} refers to an A/B test conducted across a segment of the YouTube user base, where each experiment arm typically comprises tens of millions of users. A \textit{launch} refers to a change in the algorithm deployed to the entire user base (billions of users). \textit{Offline analysis} refers to model performance metrics (e.g., AUC) calculated on a held-out evaluation set after training.

\textbf{Dataset Statistics:} Billions or more of training examples are generated daily. Each model consumes a time window spanning several weeks of data. The evaluation set comprises 1\%.

\textbf{Empirical Evaluation Context:} First, while some reported metric changes may appear small in magnitude (e.g., 0.1\%), they translate to substantial and statistically significant absolute impacts given the scale of the user base. Furthermore, such incremental gains are meaningful for mature and already optimized systems. Second, the proprietary nature of the production environment means some of the precise mathematical formulations for our experiment metrics and prediction labels are necessarily abstracted, which necessitates a certain degree of assumed domain knowledge. We acknowledge that these specific design choices and metric definitions will vary across other applications.

\textbf{Discussion of Results:} Offline AUC evaluation yields generally positive but mixed results; while some original heads show improvements, others remain neutral, with occasional minor regressions in unrelated heads. However, all live experiments demonstrate statistically significant gains in both user engagement and satisfaction. This confirms an expansion of the system’s operating point rather than a trade-off between objectives. The metrics are selected to illustrate this duality, such as Daily Active Users and Opt-outs for notifications (Sec.~\ref{sec:notif_initial}).

\begin{table*}[htbp]
    \caption{Summary of Results. \textmd{Note: Reported metrics may represent proprietary variants of the listed descriptors.}}
    \label{tab:results_summary}
    \centering
    \setlength{\tabcolsep}{3pt}
    \begin{tabular}{@{} >{\raggedright\arraybackslash}p{0.10\linewidth} >{\raggedright\arraybackslash}p{0.14\linewidth} >{\raggedright\arraybackslash}p{0.22\linewidth} >{\raggedright\arraybackslash}p{0.18\linewidth} >{\raggedright\arraybackslash}p{0.31\linewidth} @{}}
        \toprule
        \textbf{Surface} & \textbf{Hypothesis} & \textbf{Auxiliary Heads} & \textbf{AUC Impact} & \textbf{Empirical Impact [95\% CI]} \\
        \midrule
        
        \textbf{Notifications} \newline (Sec.~\ref{sec:notif_initial}) & Combined: Transfer Learning + RL Input & \textbf{10 heads ($D \times V$):} Cross-labels between direct actions $D$ (Open/Dismissal) and ambient activity $V$ (Visits). & $+$3--8\% (Direct $D$); \newline Neutral (Ambient $V$) & $+$0.17\% {\scriptsize [$+$0.01, $+$0.32]} Daily Active Users \newline $+$0.19\% {\scriptsize [$+$0.02, $+$0.37]} Valued watch time \newline $-$1.05\% {\scriptsize [$-$1.84, $-$0.26]} Opt-outs \newline $-$7.21\% {\scriptsize [$-$7.30, $-$7.11]} Sends \newline $+$7.67\% {\scriptsize [$+$7.34, $+$7.99]} Click-Through Rate \\
        \midrule
        
        \textbf{Homepage} \newline \textit{Phase 1} \newline (Sec.~\ref{sec:home_multi}) & Transfer Learning & \textbf{5 unconditioned heads:} \newline Watch Time Ratio, Save, Like, Subscribe, Inline Playback. & $+$0.2--0.5\% (Primary: \textit{click}, \textit{subscribe}, \textit{inline playback}, \textit{like}, \textit{save}); \newline Minor regressions (Unrelated: \textit{scroll}, \textit{share}) & $+$0.05\% {\scriptsize [$+$0.01, $+$0.09]} Valued watch time \newline $+$0.10\% {\scriptsize [$+$0.05, $+$0.16]} Shorts engagement \newline $-$0.24\% {\scriptsize [$-$0.42, $-$0.05]} Low quality impressions \\
        \midrule
        
        \textit{Phase 2} \newline (Sec.~\ref{sec:home_multi}) & Transfer Learning & \textbf{7 heads:} Share (uncond.), Download (cond.), plus \textbf{5 heads:} Save $\times$ [Action] conditioned. & $+$1--2\% (\textit{share}, \textit{download}); \newline $\approx -$0.5\% (\textit{like}, \textit{dislike}) & $+$0.05\% {\scriptsize [$+$0.00, $+$0.10]} Valued watch time \newline $+$0.06\% {\scriptsize [$+$0.00, $+$0.12]} Shorts engagement \newline $-$0.16\% {\scriptsize [$-$0.30, $-$0.03]} Low quality impressions \\
        \midrule
        
        \textit{Phase 3} \newline (Sec.~\ref{sec:home_rl}) & RL Input & \textbf{11 heads:} Final refined set of unconditioned, conditioned, and cross-task predictions. & N/A; baseline contains MTL changes & $+$0.03\% {\scriptsize [$+$0.00, $+$0.07]} Home engagement \newline $+$0.10\% {\scriptsize [$+$0.05, $+$0.16]} Shorts engagement \newline $-$0.64\% {\scriptsize [$-$0.86, $-$0.43]} Low quality impressions \\
        \midrule
        
        \textbf{Watch Next} \newline (Sec.~\ref{sec:watch_next}) & Transfer Learning & \textbf{4 unconditioned variants:} \newline Continuation, Like, Dislike, Subscribe. & $+$0.2--2.6\% (Original heads) \newline $+$0.25\% (CTR), $+$0.22\% (Bad Watch) & $+$0.19\% {\scriptsize [$+$0.15, $+$0.23]} Watch Next Valued Watch Time \newline $+$0.08\% {\scriptsize [$+$0.04, $+$0.12]} Valued watch time \newline $-$0.42\% {\scriptsize [$-$0.53, $-$0.31]} Low quality impressions \\
        \bottomrule
    \end{tabular}
\end{table*}

\subsection{Notifications}
The notifications product presents unique challenges for both modeling and live experimentation. One challenge is attribution: positive signals of recommendation quality, such as visiting the app after a send, can often only be indirectly attributed to the notification. Another challenge is user state: unlike in-app recommendations served to users who have already proactively opened the app, notifications reach users at varying levels of availability and responsiveness. Consequently, baseline AUC scores for prediction tasks are often lower, and live experiments frequently exhibit wider confidence intervals. However, offline analysis can still verify accuracy improvements, and strategies such as repeating experiments or bundling changes can help achieve statistically significant improvements in quality metrics or increase confidence in the results.

\subsubsection{Initial Launch of 10 Heads}\label{sec:notif_initial}

We conducted an initial launch adding ten auxiliary heads to the Notifications user engagement multi-task model, adopting the architecture shown in Figure~\ref{fig:cross_heads_through_existing} (connecting to existing heads' top hidden layers). The added heads predicted cross-task outcomes $XY$, where $X \in D$ and $Y \in V$ are drawn from two distinct sets of existing predictions defined below:

\begin{itemize}
    \item $D$: Five predictions of notification interactions: \textit{Open}, \textit{Open (variant)}, \textit{Open (variant II)}, \textit{Dismissal}, \textit{Dismissal (variant)}.
    \item $V$: Two predictions of ambient activity indirectly attributed to the notification: \textit{Visit}, \textit{Visit (variant)}.
\end{itemize}

The live experiment utilized a three-arm design to separate the testing of Hypotheses~\ref{Transfer Learning Hypothesis} and \ref{Environment Modeling Hypothesis}: a control arm, a transfer learning arm (training only), and a combined arm that also served these predictions to the RL policy. As summarized in Table~\ref{tab:results_summary}, offline analysis confirmed accuracy improvements (primarily in direct interaction heads $D$), while the live combined arm demonstrated broad improvements across user engagement, satisfaction, and efficiency.

\subsubsection{Follow-Up Launch on More Heads}
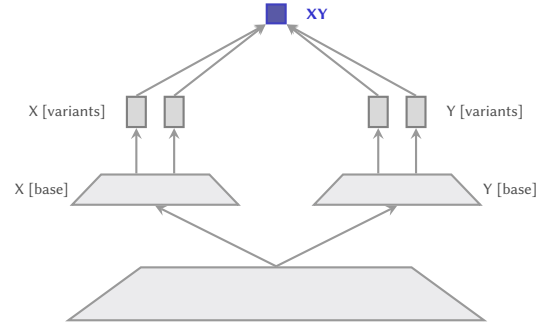
\begin{figure}[htpb]
    \centering
    \begin{tikzpicture}
        \node[nnbase, minimum width=5.5cm] (base) {};
        \node[nneventbase, above=0.8cm of base, xshift=-1.6cm] (xbase) {};
        \node[nneventbase, above=0.8cm of base, xshift=1.6cm] (ybase) {};
        
        \draw[arrow] (base.north) -- (xbase.south);
        \draw[arrow] (base.north) -- (ybase.south);
        
        \node[lbl, left=0.1cm of xbase] {X [base]};
        \node[lbl, right=0.1cm of ybase] {Y [base]};
        
        \node[nntower, above=0.6cm of xbase, xshift=-0.25cm] (x1) {};
        \node[nntower, above=0.6cm of xbase, xshift=0.25cm] (x2) {};
        \node[nntower, above=0.6cm of ybase, xshift=-0.25cm] (y1) {};
        \node[nntower, above=0.6cm of ybase, xshift=0.25cm] (y2) {};
        
        \draw[arrow] (xbase.north -| x1.south) -- (x1.south);
        \draw[arrow] (xbase.north -| x2.south) -- (x2.south);
        \draw[arrow] (ybase.north -| y1.south) -- (y1.south);
        \draw[arrow] (ybase.north -| y2.south) -- (y2.south);
        
        \node[lbl, left=0.15cm of x1] {X [variants]};
        \node[lbl, right=0.15cm of y2] {Y [variants]};
        
        \node[nnheadxy, above=3.2cm of base.north] (hxy) {};
        \node[lblnew, right=0.15cm of hxy] {XY};
        
        \draw[arrow] (x1.north) -- (hxy.south west);
        \draw[arrow] (x2.north) -- (hxy.south west);
        \draw[arrow] (y1.north) -- (hxy.south east);
        \draw[arrow] (y2.north) -- (hxy.south east);
    \end{tikzpicture}
    \caption{Gradients flow through variant-specific towers.}
    \label{fig:cross_heads_on_variants}
    \Description{Neural network diagram illustrating gradients flowing from cross-task heads.}
\end{figure}

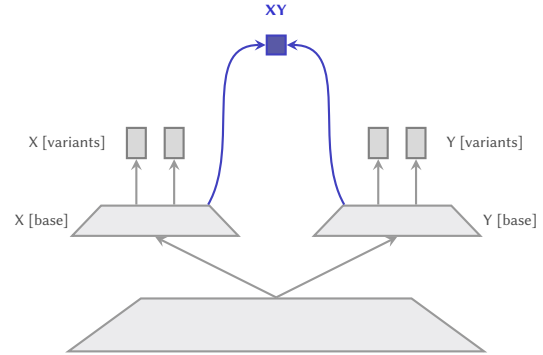
\begin{figure}[htpb]
    \centering
    \begin{tikzpicture}
        \node[nnbase, minimum width=5.5cm] (base) {};
        \node[nneventbase, above=0.8cm of base, xshift=-1.6cm] (xbase) {};
        \node[nneventbase, above=0.8cm of base, xshift=1.6cm] (ybase) {};
        
        \draw[arrow] (base.north) -- (xbase.south);
        \draw[arrow] (base.north) -- (ybase.south);
        
        \node[lbl, left=0.1cm of xbase] {X [base]};
        \node[lbl, right=0.1cm of ybase] {Y [base]};
        
        \node[nntower, above=0.6cm of xbase, xshift=-0.25cm] (x1) {};
        \node[nntower, above=0.6cm of xbase, xshift=0.25cm] (x2) {};
        \node[nntower, above=0.6cm of ybase, xshift=-0.25cm] (y1) {};
        \node[nntower, above=0.6cm of ybase, xshift=0.25cm] (y2) {};
        
        \draw[arrow] (xbase.north -| x1.south) -- (x1.south);
        \draw[arrow] (xbase.north -| x2.south) -- (x2.south);
        \draw[arrow] (ybase.north -| y1.south) -- (y1.south);
        \draw[arrow] (ybase.north -| y2.south) -- (y2.south);
        
        \node[lbl, left=0.15cm of x1] {X [variants]};
        \node[lbl, right=0.15cm of y2] {Y [variants]};
        
        \node[nnheadxy, above=3.2cm of base.north] (hxy) {};
        \node[lblnew, above=0.15cm of hxy] {XY};
        
        \draw[arrownew] ([xshift=0.7cm]xbase.north) to[out=60, in=180] (hxy.west);
        \draw[arrownew] ([xshift=-0.7cm]ybase.north) to[out=120, in=0] (hxy.east);
    \end{tikzpicture}
    \caption{Gradients flow directly to shared event base layers.}
    \label{fig:cross_heads_on_event_base}
    \Description{Neural network diagram illustrating gradients flowing directly bypassing variants.}
\end{figure}

A subsequent deployment replaced the existing independent, overlapping variants in $D$ with a set of disjoint variants sharing a common base layer. This architectural shift raised a critical question regarding the optimal gradient path for learning pre-existing cross-task relationships. In the newer architecture, the number of potential cross-task heads scales with the number of variants modeled per event, necessitating a re-evaluation of how gradients from the cross-task loss are backpropagated to the shared body for transfer learning. We compared two distinct gradient pathways:

\begin{itemize}[leftmargin=*]
    \item \textbf{Path 1: Connect to Variant Layers.} Gradients flow from the cross-task heads through the specialized (variant) tower layers, see Figure~\ref{fig:cross_heads_on_variants}.

    \item \textbf{Path 2: Connect Directly to Shared Base.} Gradients flow directly from the cross-task heads to the shared base layer, bypassing the variant-specific towers, see Figure~\ref{fig:cross_heads_on_event_base}.

\end{itemize}

Empirical results demonstrated that Path 2 significantly outperformed Path 1, particularly on class-imbalanced and sparse-label tasks, where AUC performance consistently improved by approximately $0.6\%\text{--}4\%$. This strongly suggests that direct regularization of the event base layers via the cross-task loss yields a higher-quality, non-specialized shared representation. We hypothesize that the direct connection in Path 2 prevents cross-task gradients from being filtered or attenuated by intermediate, variant-specific tower layers, which are otherwise optimized to discard information irrelevant to their specific variant. We plan to conduct further research on optimizing gradient paths for auxiliary heads.

\subsection{Homepage}

The Homepage serves as the primary landing surface for users initiating their viewing journey. Unlike the Notifications surface, the Homepage has high intent since users have actively decided to open the app. This typically yields cleaner signals and more statistically significant results in live experiments. The underlying multi-task model is also significantly larger with a high number of existing heads, which creates a larger combinatorial space of potential task pairs for relationship modeling. Consequently, a challenge is the strategic selection and refinement of the auxiliary heads. The model architecture is centered around one prediction of critical importance, $P(\text{click})$, with numerous heads conditioned on positive click labels.

We ultimately utilized a combination of the auxiliary head types described in Section~\ref{Modeling: Estimates of the Joint Distribution}, including both unconditioned and cross-label heads.

\subsubsection{Multi-Task Model Auxiliary Heads}\label{sec:home_multi}

We conducted two primary launches introducing auxiliary heads, followed by a cleanup phase. The auxiliary heads were connected directly to the shared body, following the architecture in Figure~\ref{fig:cross_heads_on_shared_body}. Both launches yielded significant improvements in user metrics, even when the heads were initially added as passive tasks (not served) to leverage transfer learning (Section~\ref{Transfer Learning Hypothesis}). Notably, the transfer learning effect was not uniformly positive; some existing heads experienced minor regressions, though these were outweighed by gains in other areas. However, live results demonstrated a net positive impact on user experience. 

\textbf{First Main Launch:} We added five auxiliary heads, unconditioned on click:
\begin{itemize}
    \item \textit{Watch Time Ratio} (uncond.) (A regression head but a derivation similar to Section~\ref{Deriving Covariance} applies to model the covariance between the binary click and continuous watch time ratio).
    \item \textit{Save}, \textit{Like}, \textit{Subscribe}, and \textit{Inline Playback} (all uncond.).
\end{itemize}

There were modest and heterogeneous improvements in the AUC scores of existing tasks, and live experiments showed substantial improvements in user engagement and quality metrics (Table~\ref{tab:results_summary}). 

\textbf{Second Main Launch:} We added seven passive auxiliary heads. They included two that reversed the conditioning of existing tasks:
\begin{itemize}
    \item \textit{Share} (uncond.).
    \item \textit{Download} \textbf{cond.} (existing task was uncond.).
\end{itemize}
The remainder were five cross-label tasks conditioned on click:
\begin{itemize}
    \item \textit{Save} $\times$ \textit{Action}, where \textit{Action} is one of: \textit{Share}, \textit{Download}, \textit{Channel Engagement}, \textit{Homepage Engagement}, \textit{Like} (all cond.).
\end{itemize}

Again, there was a modest and heterogeneous AUC impact, and positive shifts in live user metrics (Table~\ref{tab:results_summary}).

\subsubsection{Passing Heads to RL Policy Model}\label{sec:home_rl}

Following an iterative selection process, we finalized a collection of 11 cross-task relation heads within the Homepage multi-task user engagement model.

In this phase, we incorporated the predictions from all 11 heads as input features to the Homepage's on-policy reinforcement learning (RL) ranking model. The control arm retained these 11 heads in the upstream multi-task user engagement model but suppressed their outputs from the learned ranking model. Since both arms benefited equally from transfer learning, this design allowed us to isolate the specific utility of the cross-task predictions in providing a richer representation of user intent, as proposed in Hypothesis~\ref{Environment Modeling Hypothesis}. 

The final set of 11 heads included:
\begin{itemize}
    \item Three head variants with click conditioning: \textit{Download} (cond.), \textit{Save} (uncond.), \textit{Share} (uncond.).
    \item Five heads of the form \textit{Save} $\times$ \textit{Action} (cond.) where \textit{Action} is one of: \textit{Share}, \textit{Download}, \textit{Channel Engagement}, \textit{Homepage Engagement}, \textit{Like}.
    \item Three unconditioned cross-label heads: \textit{Save} $\times$ \textit{Share}, \textit{Save} $\times$ \textit{Download}, \textit{Share} $\times$ \textit{Download} (all uncond.).
\end{itemize}

The live experiment yielded positive results on user engagement quality, though the magnitude of impact was smaller than the gains observed from the initial transfer learning experiments (Table~\ref{tab:results_summary}). The final refinement and full-scale deployment are in progress.

\subsection{Watch Next}

The Watch Next system recommends content within the video watch page environment. It leverages the current video as its primary signal context and must accommodate diverse user journey entry points. A key challenge is quantifying the dynamics between Watch Next and other YouTube surfaces such as the Homepage and the Shorts Infinite Player. To address this, impact is measured across all of YouTube instead of just the Watch Next surface. At the moment, only Phase 1, assessing transfer learning benefits (see Section~\ref{Transfer Learning Hypothesis}), is complete.

\subsubsection{Addition of Four Auxiliary Non-Serving Heads}\label{sec:watch_next}

The launch on the Watch Next sidebar introduced variants based on heads previously conditioned on a watch requirement. These heads were connected directly to the shared body, as in Figure~\ref{fig:cross_heads_on_shared_body}.

\begin{itemize}
    \item \textit{Continuation}, \textit{Like}, \textit{Dislike}, \textit{Subscribe} (all uncond.).
\end{itemize}

The treatment model yielded consistent offline AUC improvements across the original serving heads and translated into significant live gains (Table~\ref{tab:results_summary}).

\section{Discussion}

\subsection{Contrasting with Scalar Loss Weighting}

A common question is whether similar improvements could be achieved by increasing the loss weights of specific heads, rather than adding an auxiliary task to predict cross-task relationships. In response, we compared adding auxiliary heads against duplicating existing heads or increasing the loss weights of existing heads, and we found no experimental evidence of improvement. Furthermore, increasing loss weights creates a fundamentally different learning dynamic than adding cross-task correlation heads:

\begin{itemize}
    \item When existing heads predict labels $X$ and $Y$ and the added head predicts $XY$, increasing the loss weight for either $X$ or $Y$ uniformly increases the training importance on all examples positive for that label. In contrast, adding a head predicting $XY$ concentrates the gradient signal specifically on the subset of examples where both tasks are positive. This subset of examples can be small if both $X$ and $Y$ are relatively sparse.
    \item When existing heads predict $X$ and $Y \mid X$ (e.g., when $X=\text{click}$), adding a head for the unconditional prediction of $Y$ allows the model to train on examples where $Y$ is positive but $X$ is not (even if they are rare). This learning signal cannot be replicated by adjusting the loss weight of $Y \mid X$, as the conditional head does not train on negative $X$ examples.
\end{itemize}

\subsection{Difference Between Crossing Task Labels and Crossing Existing Task Predictions}

Suppose the policy model consuming the multi-task predictions is a neural network capable of functionally approximating feature crosses \citep{hornik1989multilayer}, potentially augmented by a specialized layer such as a Deep Cross Network (DCN) \citep{wang2017deep}. A frequently asked question is: What is the added value of explicitly predicting the cross-label probability $\mathbb{E}[XY]$ when $\mathbb{E}[X]$ and $\mathbb{E}[Y]$ are already present in the policy model's feature set?

The answer is that while the product $\mathbb{E}[X]\mathbb{E}[Y]$ can be approximated, $\mathbb{E}[X]\mathbb{E}[Y] \neq \mathbb{E}[XY]$ in general. The motivation for learning cross-task relationships is precisely this inequality. The difference between the two terms is the covariance, as detailed in Section~\ref{Deriving Covariance}. The limiting factor is not the sophistication of functional transformations applied to $\mathbb{E}[X]$ and $\mathbb{E}[Y]$, but rather the availability of information in the feature set; the joint distribution inherently contains greater information than the marginals, as explained in Section~\ref{Marginal and Joint Distributions}.

\section{Conclusion and Future Work}

We proposed a framework extending multi-task learning to explicitly model or approximate the joint distribution between tasks, moving beyond the standard practice of relying solely on joint training for implicit feature sharing. We formulated a practical solution by introducing auxiliary heads to approximate covariance and pairwise joint distributions, demonstrating significant positive impacts on user engagement through large-scale deployments in YouTube's recommender systems.

Future work can focus on two primary directions. The first is a principled, automated approach for selecting auxiliary heads, reducing the current reliance on empirical trial-and-error. The second is a structural evolution toward merging the User Engagement Model and the RL policy into a unified, end-to-end architecture---similar to paradigms found in autonomous driving \citep{codevilla2018end}. Such a unification would allow the decision policy to access the totality of learned task relationships directly from the latent representation, potentially obviating the need for explicit auxiliary heads and inviting an alternative formulation of learned task relationships.

\begin{acks}
We are grateful for the insightful discussions and code reviews provided by colleagues not listed as authors, including Shawn Andrews, Kris Peterson, Yash Vadalia, and Angela Yan. We also extend our thanks to the broader Notifications, Home, and Watch Next teams at YouTube for their support.
\end{acks}

\balance
\bibliographystyle{ACM-Reference-Format}
\bibliography{references}

\end{document}